\documentclass[10pt,twocolumn,letterpaper]{article}

\usepackage{cvpr}              
\definecolor{cvprblue}{rgb}{0.21,0.49,0.74}
\usepackage[pagebackref,breaklinks,colorlinks,allcolors=cvprblue]{hyperref}
\usepackage[accsupp]{axessibility}  
\usepackage{rotating}
\usepackage{tabularx}
\usepackage{adjustbox}
\usepackage{multirow}
\usepackage[table,xcdraw]{xcolor}
\usepackage{booktabs}
\usepackage[normalem]{ulem}
\useunder{\uline}{\ul}{}

\def\paperID{44982} 
\def\confName{CVPR}
\def\confYear{2026}

\title{Language-Grounded Decoupled Action Representation for Robotic Manipulation}

\author{
Wuding Weng, \quad 
Tongshu Wu, \quad 
Liucheng Chen, \quad 
Siyu Xie, \quad
Zheng Wang\thanks{Corresponding author.}, \\[1mm] 
Xing Xu, \quad 
Jingkuan Song, \quad 
Heng Tao Shen \\[1mm]
School of Computer Science and Technology, Tongji University, China \\
{\tt\small  55dupup@gmail.com, zh\_wang@hotmail.com}
}

\begin{document}
\maketitle
\begin{abstract}
The heterogeneity between high-level vision-language understanding and low-level action control remains a fundamental challenge in robotic manipulation. Although recent methods have advanced task-specific action alignment, they often struggle to generate robust and accurate actions for novel or semantically related tasks. To address this, we propose the \textbf{La}nguage-Grounded \textbf{D}ecoupled \textbf{A}ction Representation (\textbf{LaDA}) framework, which leverages natural language as a semantic bridge to connect perception and control. LaDA introduces a fine-grained intermediate layer of three interpretable action primitives—translation, rotation, and gripper control—providing explicit semantic structure for low-level actions. It further employs a semantic-guided soft-label contrastive learning objective to align similar action primitives across tasks, enhancing generalization and motion consistency. An adaptive weighting strategy, inspired by curriculum learning, dynamically balances contrastive and imitation objectives for stable and effective training. Extensive experiments on simulated benchmarks (LIBERO and MimicGen) and real-world demonstrations validate that LaDA achieves strong performance and generalizes effectively to unseen or related tasks.

\end{abstract}    
\section{Introduction}

Achieving efficient and generalizable alignment among visual, linguistic, and action representations remains a fundamental challenge in robotic manipulation~\cite{zhao2023learning,liu-niu2025lbp,fan2025interleave,wu2025momanipvla,lv2025spatial}. 
Recent advances in vision-language-action (VLA) models~\cite{li2025bridgevla,li2024cogact,zhong2025dexgraspvla,wen2025dexvla,wen2025diffusionvla} have demonstrated the potential of mapping multimodal observations directly to executable control actions. 
However, a persistent gap remains between high-level semantic instructions and fine-grained motor execution. 
Semantically distinct tasks such as “pour water” and “place bottle” often share underlying motion primitives—e.g., reaching, grasping, and rotating—yet current models fail to exploit these shared components, resulting in redundant learning and poor cross-task generalization.

\begin{figure}[t]
  \centering
  \includegraphics[width=\linewidth]{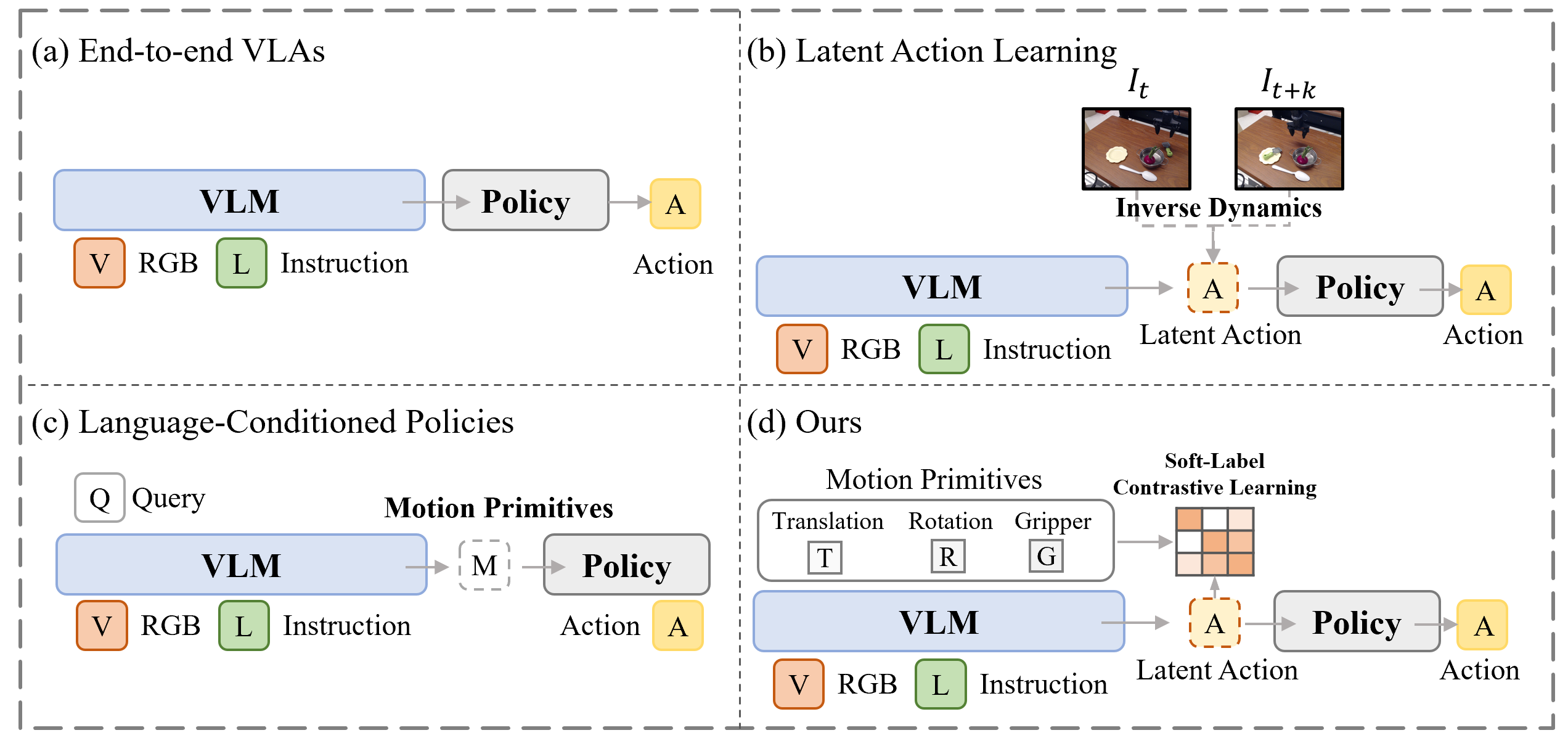}
   \caption{
   Comparison of representative paradigms in vision-language-action learning. Existing approaches either entangle perception and control in end-to-end VLAs, rely on latent action embeddings without explicit semantics, or use discrete language-conditioned primitives that lack fine-grained motion grounding. Our LaDA framework bridges this gap by leveraging language as a semantic bridge to decouple and align vision, language, and action representations through soft-label contrastive learning, enabling semantically grounded and generalizable manipulation.
}
   \label{fig1.existing methods}
\end{figure}

As illustrated in \cref{fig1.existing methods}, existing approaches for connecting vision, language, and action can be broadly categorized into three paradigms. 
(a) Vision-Language-Action (VLA) models \cite{reed2022generalist,brohan2022rt,ghosh2024octo,kim2025openvla,li2024vision} learn direct end-to-end mappings from multimodal inputs to low-level control through unified transformer-based policies. 
While scalable, these architectures entangle perception and control, limiting interpretability and the reuse of shared motion structures. 
(b) Latent Action Learning \cite{kim2025uniskill,schmidtlearning,yelatent,huvideo} encodes actions into compact latent spaces, abstracting dynamics between visual observations. 
However, these latent embeddings are typically defined by observation deltas without explicit semantics, hindering cross-task transferability. 
(c) Language-Conditioned Policies \cite{yao2025think,xia2025phoenix,kang2024cliprt} introduce natural language as task supervision or intermediate representation, enhancing interpretability but relying on coarse, discrete primitives (e.g., “move forward”, “close gripper”) that fail to capture fine-grained motion parameters such as translation magnitude or rotation axis. Consequently, existing paradigms achieve either semantic understanding or precise control—but rarely both—leaving an open question: \textit{How can we construct an action representation that is both semantically grounded and transferable across tasks?}

We posit that the root cause of this misalignment lies in the absence of a semantic grounding layer bridging symbolic intent and continuous execution. \textit{Language} naturally serves this role—it provides a universal interface connecting human intent, visual perception, and robotic control \cite{belkhale2024rt,jiang2025multimodal}. Unlike purely visual or kinematic representations \cite{du2023learning,yelatent}, linguistic abstraction encodes compositional and semantic regularities, offering a shared space where motion concepts can be compared, transferred, and generalized. By representing low-level actions through language-grounded primitives, we can endow control trajectories with explicit semantics, enabling consistent alignment between visual, linguistic, and motor representations under a unified interpretive framework. 

We propose the \textbf{Language-Grounded Decoupled Action Representation (LaDA)} framework, which leverages language as a semantic bridge to unify high-level visual–linguistic understanding with low-level control. LaDA introduces a fine-grained intermediate layer consisting of three interpretable motion primitives—translation, rotation, and gripper control—each associated with a natural-language description. This decomposition provides explicit semantic supervision for low-level actions and exposes shared motion structures across tasks.
Built upon this semantic abstraction, LaDA employs a \textit{semantic-guided soft-label contrastive objective} that assigns continuous affinity weights among action descriptions, promoting mutual reinforcement of semantically related motions within a shared embedding space.
Finally, an \textit{adaptive weighting mechanism}, inspired by curriculum learning \cite{wang2021survey}, dynamically balances contrastive and imitation objectives, ensuring stable convergence and effective semantic alignment.

Our main contributions are summarized as follows:
\begin{itemize}
    \item We present LaDA, a unified framework that bridges high-level vision–language understanding and low-level control by decomposing continuous 7-DoF actions into interpretable, language-grounded primitives—translation, rotation, and gripper control. This design provides explicit semantic supervision for low-level motion and enables fine-grained cross-task alignment and compositional generalization.

    \item We develop a semantic-guided soft-label contrastive objective that captures continuous affinities among motion primitives and integrates an adaptive weighting mechanism to balance contrastive and imitation objectives. This formulation ensures stable optimization and progressive refinement of motion semantics across tasks.
    \item We extensively evaluate LaDA across both simulated and real-world robotic manipulation benchmarks, including LIBERO and MimicGEN, demonstrating state-of-the-art performance and strong generalization to unseen and semantically related tasks.
\end{itemize}

\section{Related Work}
\label{sec: related work}

\noindent\textbf{Vision-Language-Action Models.} Recent advances in vision–language–action (VLA) models have driven progress in multimodal robotic learning by mapping high-dimensional sensory inputs to low-level motor actions \cite{zhou2025mitigating,pan2025omnimanip,li2025object,ji2025robobrain,lin2025showui,huang2025roboground}. Early generalist systems such as Gato \cite{reed2022generalist}, RT-1 \cite{brohan2022rt}, and Octo \cite{ghosh2024octo} demonstrated the scalability of transformer-based architectures for learning shared policies across tasks, embodiments, and environments. Subsequent works, including RT-2 \cite{zitkovich2023rt}, OpenVLA \cite{kim2025openvla}, and RoboFlamingo \cite{li2024vision}, further integrated visual grounding and linguistic reasoning by treating robotic actions as language tokens or augmenting pretrained vision–language backbones with action decoders. However, despite their success in multimodal understanding \cite{11396092,11358795,11048677,10507035}, these end-to-end models tightly couple perception and control, lacking explicit structural disentanglement. As a result, they fail to capture reusable motion semantics across tasks, leading to redundant learning and limited generalization.

\begin{figure*}[t]
  \centering
  \includegraphics[width=\linewidth]{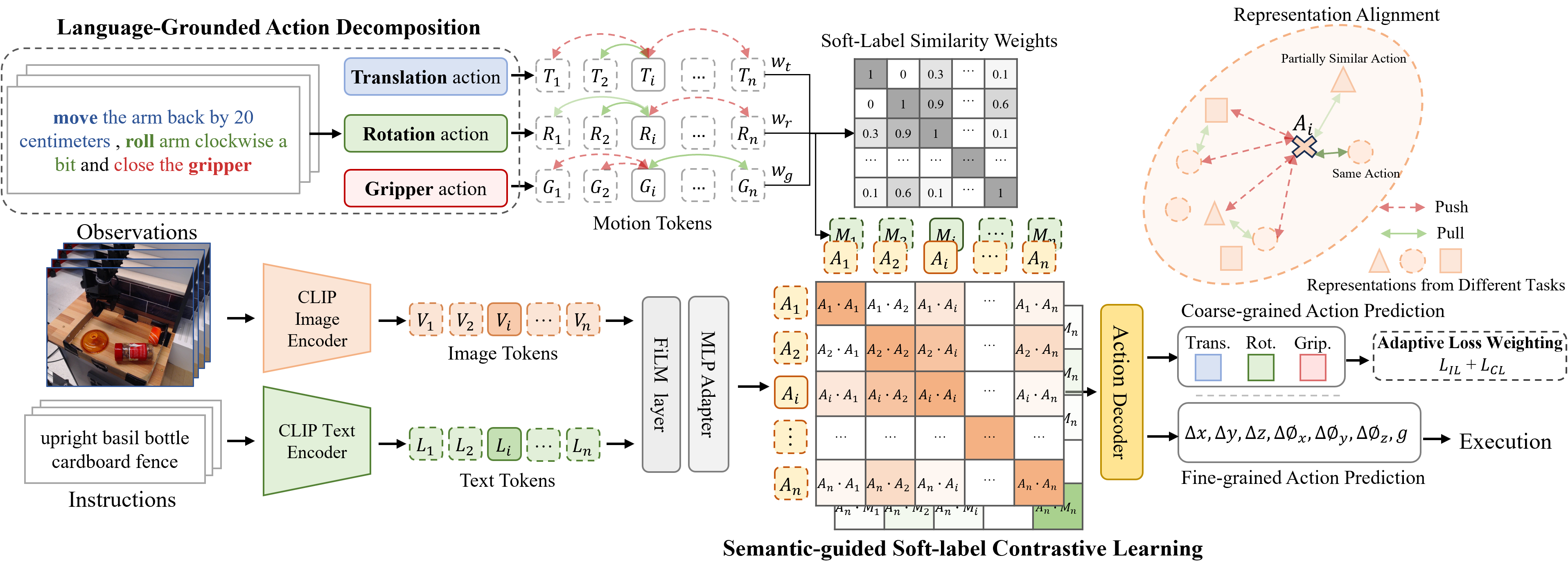}
   \caption{
    Overview of the proposed \textbf{LaDA} framework. LaDA leverages language as a semantic bridge to connect high-level vision–language understanding with low-level control. It decomposes continuous 7-DoF end-effector actions into interpretable primitives—translation, rotation, and gripper control—and encodes them within a shared semantic embedding space. Semantic-guided soft-label contrastive learning aligns multimodal representations across tasks, while an adaptive weighting strategy dynamically balances imitation and contrastive objectives, enabling efficient cross-task transfer and robust generalization.
   }

   \label{overview}
\end{figure*}

\noindent\textbf{Latent Action Learning.}
To reduce the complexity of direct perception-to-control mapping, another research line focuses on latent action learning, which seeks compact, robot-agnostic representations that mediate between perception and dynamics. Methods such as Genie \cite{bruce2024genie}, Dynamo \cite{cui2024dynamo}, and R3M \cite{nairr3m} pretrain general-purpose visual encoders on large-scale video corpora to support downstream policy learning. Building on this foundation, latent action models—UniSkill \cite{kim2025uniskill}, LAPO \cite{schmidtlearning}, and LAPA \cite{yelatent}—jointly train forward and inverse dynamics models to infer latent representations that capture transition intent between observation pairs, even without explicit action labels. Further extensions such as UniAct \cite{zheng2025universal} and UniVLA \cite{bu2025univla} aim to enhance cross-embodiment or task-centric transferability. Nonetheless, most latent spaces are still defined by visual deltas rather than semantically grounded motion concepts, restricting their ability to represent shared motion structures essential for generalizable manipulation.

\noindent\textbf{Language-Conditioned Policies.}
Motivated by the semantic limitations of latent actions, recent studies have explored using language as an intermediate representation to bridge symbolic intent and continuous control. Frameworks such as SayCan \cite{ahn2022can} and CLIPORT \cite{shridhar2022cliport} ground natural-language commands into executable policies by combining pretrained vision–language models with reinforcement or imitation learning. While these approaches improve interpretability, they provide weak supervision over fine-grained motion parameters. More recent methods introduce language into the action space itself: RT-H \cite{belkhale2024rt} and Phoenix \cite{xia2025phoenix} jointly predict linguistic motion descriptions and corresponding trajectories for online correction, whereas CLIP-RT \cite{kang2024cliprt} formulates control as classification over discrete language-based motion tokens. PPL \cite{yao2025think} further proposes reusable primitive prompts for compositional and lifelong learning. Despite these advances, the primitives used remain coarse and under-parameterized—typically lacking explicit spatial attributes such as translation magnitude or rotation angle—thus failing to achieve precise multimodal alignment. In contrast, our proposed LaDA framework introduces a language-grounded, decoupled representation that explicitly parameterizes these primitives. By aligning them via soft-label contrastive learning, LaDA provides interpretable, semantically consistent supervision for continuous control, bridging the gap between high-level visual–linguistic reasoning and low-level motor execution.

\section{Approach}
\subsection{Overview}
A central challenge in vision–language–action learning is constructing an action representation that links high-level semantics to low-level control while remaining transferable across tasks. Conventional approaches either encode actions directly in continuous 7-DoF space or rely on coarse, predefined primitives, making it difficult to capture fine-grained motion semantics that generalize across diverse trajectories and task contexts.

To address this limitation, we introduce LaDA, a framework that uses language as a semantic bridge to unify vision, language, and action within a shared embedding space. As illustrated in \cref{overview}, LaDA first decomposes each continuous action into three interpretable, language-grounded primitives—translation, rotation, and gripper state—providing explicit semantic structure for downstream alignment. Built on this representation, LaDA applies a semantic-guided soft-label contrastive objective that aligns actions according to their primitive-level semantic affinity, and incorporates an adaptive loss weighting strategy to balance imitation-based supervision with fine-grained semantic alignment throughout training. Together, these components produce an interpretable and transferable action space that supports efficient cross-task knowledge sharing and strong compositional generalization.

\subsection{Language-Grounded Action Decomposition}
\label{subsec decomposition}

To introduce interpretable structure into continuous robot control and facilitate transferable skill learning, LaDA decomposes each 7-DoF end-effector action $\mathbf{a}_t$ into a set of orthogonal and language-grounded motion primitives, each corresponding to a distinct dimension of control. Formally, we define a projection $\Pi: \mathbf{a}_t \mapsto \mathbf{p}_t$, yielding three primitives that represent fundamental motion components: 
(1) Translation Primitive $(\Delta T)$ — expressed through linguistic templates such as “Move [dist] meters along [dir]”;  
(2) Rotation Primitive $(\Delta R)$ — described as “Rotate [mag] degrees around [axis]”;  
(3) Gripper Primitive $(G)$ — represented by discrete commands “Open” or “Close”.  

Each primitive is discretized into symbolic, language-aligned bins, transforming continuous control trajectories into interpretable semantic categories. This decomposition bridges the gap between low-level kinematics and high-level semantics, allowing actions to be represented and compared within a shared linguistic space. By providing explicit semantic grounding for each motion component, this process establishes the foundation for cross-task action alignment and compositional understanding across diverse manipulation skills.

\subsection{Semantic-Guided Contrastive Learning}
Building on this decomposition, LaDA leverages language as a semantic scaffold to align multi-modal embeddings of vision, language, and action. This alignment enforces semantic consistency across tasks by associating actions with similar primitive semantics, even under diverse control trajectories and task contexts.

Concretely, given visual observations $\mathit{V}_t$, language instructions $\mathit{L}_t$, and corresponding low-level actions $\mathbf{a}_t = [\Delta x, \Delta y, \Delta z, \phi_x, \phi_y, \phi_z, g]$, LaDA learns embeddings where actions with semantically related primitives—such as similar translation direction, rotation axis, or gripper state—are placed closer together. Unlike conventional contrastive learning that relies on discrete positive or negative pairs, LaDA utilizes a continuous notion of semantic affinity guided by linguistic similarity. This enables soft alignment among partially related actions, capturing fine-grained motion correspondences and preserving shared semantics across tasks. Consequently, it forms a unified latent space that supports cross-task representation sharing and generalization to novel instructions.

\subsubsection{Soft-Label Similarity Construction}
To operationalize this fine-grained alignment, LaDA constructs a soft-label similarity matrix $\mathit{S} \in [0,1]^{N \times N}$ that encodes linguistic affinities between discretized motion primitives. 
By aligning partially related actions with graded weights, $\mathit{S}$ quantitatively captures the degree of correspondence between actions that share similar translation, rotation, or gripper attributes, forming the basis for soft-label contrastive learning.

Formally, $\mathit{S}$ integrates primitive-level correspondences across translation, rotation, and gripper dimensions as:
\begin{equation}
\label{similarity matrics}
\mathit{S} = \frac{w_t \mathit{M}_t + w_r \mathit{M}_r + w_g \mathit{M}_g}{w_t + w_r + w_g},
\end{equation}
where $\mathit{M}_t$, $\mathit{M}_r$, and $\mathit{M}_g$ denote binary match matrices indicating whether two actions share the same primitive attribute, and $(w_t, w_r, w_g)$ are hyperparameters controlling the relative contribution of each component.
Each entry $\mathit{S}_{ij}$ represents a graded measure of primitive-level semantic similarity between actions $i$ and $j$, serving as fine-grained supervision for LaDA’s soft-label contrastive objective.

\subsubsection{Soft-Label Dual-Path Contrastive Learning}
\label{subsubsec:contrastive_imitation}
Building on the semantic affinities encoded in $\mathit{S}$, LaDA introduces a \textbf{Dual-Path Soft-Label Contrastive Learning} objective that jointly aligns visual, linguistic, and action representations. The goal is twofold:
(1) encourage actions that share primitive-level intent to cluster in embedding space, and
(2) ensure that each action remains grounded in its corresponding linguistic description, preserving semantic interpretability.

Given a visual observation $\mathit{V}_i$, instruction $\mathit{L}_i$, and its primitive description $\mathcal{D}(p_i)$, LaDA first extracts modality-specific embeddings using pretrained CLIP encoders~\cite{radford21a}. The visual encoder produces an image token $\mathit{v}_i = f_v(\mathit{V}_i)$, while the text encoder yields an instruction token $\mathit{l}_i = f_l(\mathit{L}_i)$. To integrate perceptual context with linguistic intent, LaDA conditions visual features on language through FiLM and projects the fused representation with a lightweight MLP adapter: $\mathit{A}_i = \text{MLP}(\text{FiLM}(\mathit{v}_i, \mathit{l}_i))$, which serves as the unified latent embedding for action alignment.

Guided by the affinity matrix $\mathit{S}$, LaDA optimizes two contrastive paths.
(i) Action–Action Alignment. This branch enforces similarity between latent actions $(\mathit{A}_i,\mathit{A}_j)$ in proportion to $\mathit{S}_{ij}$, encouraging actions that share similar primitive attributes to appear closer in embedding space. (ii) Action–Primitive Alignment. This branch anchors each latent action to the tokenized primitive description $\mathit{P}_j = f_l(\mathcal{D}(p_j))$, ensuring that the embedding space remains explicitly tied to its linguistic semantics.

Both branches use a soft-label InfoNCE objective \cite{oord2018representation} weighted by $\mathit{S}$:
\begin{equation}
\label{branch action}
\mathcal{L}_{a} = -\sum_{i=1}^N \sum_{j=1}^N \mathit{S}_{ij} 
\log \frac{\exp(\text{sim}(\mathit{A}_i, \mathit{A}_j) / \tau)}
{\sum_{k=1}^N \exp(\text{sim}(\mathit{A}_i, \mathit{A}_k) / \tau)},
\end{equation}
\begin{equation}
\label{branch motion}
\mathcal{L}_{m} = -\sum_{i=1}^N \sum_{j=1}^N \mathit{S}_{ij} 
\log \frac{\exp(\text{sim}(\mathit{A}_i, \mathit{P}_j) / \tau)}
{\sum_{k=1}^N \exp(\text{sim}(\mathit{A}_i, \mathit{P}_k) / \tau)}
\end{equation}
where $\text{sim}(\cdot, \cdot)$ denotes cosine similarity and $\tau$ is a temperature parameter.

The overall objective combines both branches:
\begin{equation}
\label{contrastive loss}
\mathcal{L}_{\text{CL}} = \mathcal{L}_{a} + \lambda \, \mathcal{L}_{m},
\end{equation}
where $\lambda$ regulates the trade-off between intra-action consistency and language-grounded anchoring. This dual-path design embeds actions into a shared space that reflects both their visual–linguistic context and their primitive-level semantics, yielding fine-grained and task-consistent alignment across modalities.

\subsubsection{Adaptive Loss Weighting} 

Alongside semantic contrastive learning, LaDA incorporates imitation loss $\mathcal{L}_{\text{IL}}$ that predicts the discretized translation, rotation, and gripper primitives defined in \cref{subsec decomposition}. This auxiliary supervision anchors the embedding space to physically meaningful motion patterns, ensuring that semantic alignment remains connected to executable control.

However, $\mathcal{L}_{\text{IL}}$ and $\mathcal{L}_{\text{CL}}$ operate at different semantic granularities and exhibit distinct convergence behaviors. The imitation loss provides coarse primitive-level guidance, while the contrastive loss refines finer semantic relationships by enforcing the soft-label affinities encoded in $\mathit{S}$. Futhermore, to prevent either signal from dominating optimization, LaDA adopts an adaptive weighting strategy based on a moving-average (MA) estimate of each loss. Let $\mathrm{MA}(\cdot)$ denote a smoothed value computed over a sliding window of recent iterations. The weights are computed as:

\begin{equation}
\begin{aligned}
\label{eq: MA}
w_{\text{IL}} &= \frac{\mathrm{MA}(\mathcal{L}_{\text{IL}})}{\mathrm{MA}(\mathcal{L}_{\text{IL}}) + \mathrm{MA}(\mathcal{L}_{\text{CL}})}, \\
w_{\text{CL}} &= \frac{\mathrm{MA}(\mathcal{L}_{\text{CL}})}{\mathrm{MA}(\mathcal{L}_{\text{IL}}) + \mathrm{MA}(\mathcal{L}_{\text{CL}})}.
\end{aligned}
\end{equation}

The final objective becomes:
\begin{equation}
\mathcal{L}_{total} = w_{\text{CL}}\mathcal{L}_{\text{CL}} + w_{\text{IL}}\mathcal{L}_{\text{IL}}.
\end{equation}

This adaptive scheme balances coarse behavioral supervision with fine-grained semantic alignment throughout training, preventing premature overfitting to imitation signals and yielding action embeddings that remain both semantically structured and behaviorally grounded.
\subsection{Fine-tuning and Inference}

After pretraining with soft-label contrastive learning, LaDA is fine-tuned with a lightweight MLP action head~\cite{kim2025openvla} to perform fine-grained 7-DoF action prediction from visual observations and language instructions. Fine-tuning uses a standard $\mathcal{L}_1$ trajectory regression loss to refine low-level control accuracy. At inference time, LaDA directly outputs continuous actions conditioned on $(V_t, L_t)$ without requiring explicit primitive labels, enabling efficient and robust end-to-end policy execution in both simulated and real-world environments.

In summary, LaDA unifies vision, language, and action representations within a shared semantic space, enabling interpretable, transferable, and generalizable robotic manipulation across diverse tasks.

\section{Experiments}

\subsection{Pretraining Datasets}
\label{sec: pretraining}
We pretrain LaDA on the Open X-Embodiment (OXE) dataset \cite{open_x_embodiment_rt_x_2023}, a large-scale collection of over one million real-world trajectories spanning 22 robot embodiments. Following standard practice in large-scale robot learning \cite{ghosh2024octo, kim2025openvla}, we use a curated subset containing roughly 22.5 million visual frames, where each low-level action is represented as a 7-DoF control vector comprising 3D translation, 3D rotation, and a binary gripper state.

To provide explicit semantic grounding for these continuous actions, we automatically derive structured language descriptions aligned with each control vector, following common practices in language-based robot supervision \cite{belkhale2024rt,kang2024cliprt}. These descriptions specify fine-grained motion attributes—e.g., “move 0.5 meters forward, rotate 90 degrees around the z-axis, and close the gripper”—and serve as auxiliary supervision during soft-label contrastive pretraining. Incorporating these linguistic cues enables LaDA to align visual, linguistic, and motor representations within a unified semantic embedding space. For more implementation details and parameter settings, please refer to the appendix.

\subsection{Experimental Setup}
We evaluate LaDA in two complementary simulated environments—LIBERO \cite{liu2023libero} and MimicGen \cite{mandlekar2023mimicgen}—to assess both semantic generalization and fine-grained control.

LIBERO is a large-scale benchmark for language-conditioned multi-task manipulation. We follow its official protocol, which includes four task suites—Spatial, Object, Goal, and Long—ranging from short-horizon spatial reasoning to complex long-horizon control. Each episode provides visual observations and natural language instructions without privileged state access. We report average success rates over 50 randomized trials per task.

MimicGen complements LIBERO by generating contact-rich demonstrations from a small number of human examples. We evaluate nine manipulation tasks, covering long-horizon assemblies (e.g., ThreePieceAssembly) and high-precision operations (e.g., Threading), each containing about 1K demonstrations paired with descriptive language annotations.

\begin{figure}[t]
  \centering
  \includegraphics[width=\linewidth]{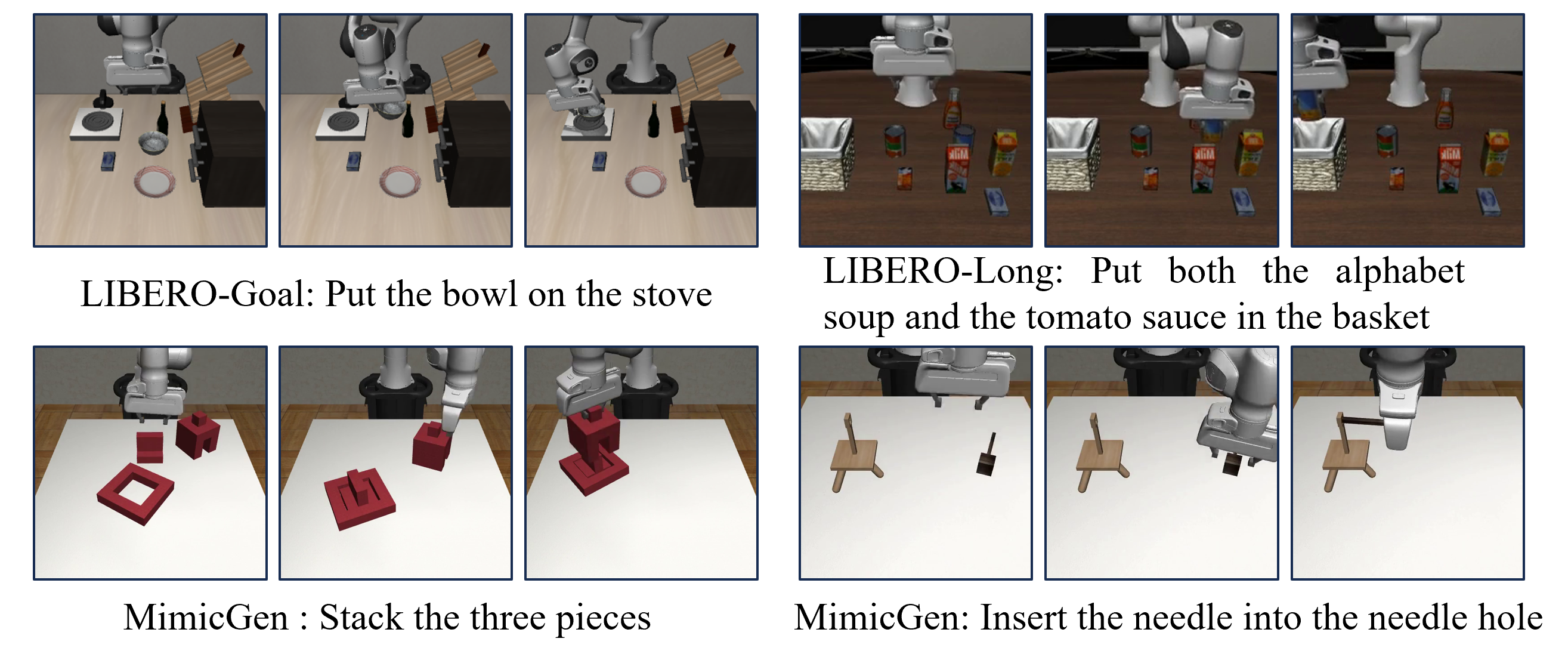}
   \caption{Example tasks from the simulation environments. (\textbf{Top}) Sample tasks from LIBERO benchmark \cite{liu2023libero}, illustrating language-conditioned manipulation scenarios. (\textbf{Bottom}) Example tasks from MimicGen \cite{mandlekar2023mimicgen}, demonstrating contact-rich manipulation skills}
   \label{fig: simulation-setup}
\end{figure}

Together, these benchmarks comprehensively evaluate LaDA’s ability to capture shared motion semantics and maintain consistent alignment 
between vision, language, and action across diverse manipulation scenarios. 
A visualization of the simulation setup is shown in \cref{fig: simulation-setup}.

\subsection{Simulation Benchmark Experiments}
\subsubsection{Experiments on LIBERO}
\noindent\textbf{Evaluation Protocol.}
We evaluate LaDA on the LIBERO benchmark \cite{liu2023libero}, which comprises four suites (Spatial, Object, Goal, and Long), each containing ten tasks with 50 human-teleoperated demonstrations. Following the standard evaluation protocol \cite{ghosh2024octo}, trajectories are preprocessed by removing idle intervals and normalizing all images to a resolution of 256×256. Models are trained on the provided demonstrations and evaluated by executing task-specific goal instructions in simulation. The overall success rate across all rollouts is reported as the primary metric.

\noindent\textbf{Baselines.}
We compare LaDA against representative VLA baselines: UniACT \cite{zheng2025universal}, LAPA \cite{yelatent}, Diffusion Policy \cite{chi2023diffusion}, Octo \cite{ghosh2024octo}, MDT \cite{reuss2024multimodal}, OpenVLA \cite{kim2025openvla}, SpatialVLA \cite{qu2025spatialvla}, CoT-VLA \cite{zhao2025cot}, WorldVLA \cite{cen2025worldvla}, Dita \cite{hou2025dita}, ThinkAct \cite{huang2025thinkact}, $\pi$-FAST \cite{pertsch2025fast}, GR00T-N1.5 \cite{bjorck2025gr00t}, MolmoAct \cite{lee2025molmoact}, FlowVLA \cite{zhong2025flowvla}, and CLIP-RT \cite{kang2024cliprt}. For fair comparison, we reimplement CLIP-RT with the same ViT-L/14 backbone as LaDA, denoted as CLIP-RT*. All models are evaluated under identical simulation and language-instruction settings, using either official hyperparameters or released checkpoints.

\noindent\textbf{Results and Discussion.}
As summarized in \cref{tab:libero_results}, LaDA attains an average success rate of 93.6\% on the LIBERO benchmark, showing consistently strong performance across all task suites. Even without trajectory augmentation and using roughly half the parameters of CLIP-RT (1.3B), LaDA performs on par or slightly better, suggesting that its language-grounded action decomposition contributes to more data-efficient learning. Compared with end-to-end VLAs (e.g., OpenVLA), latent-action VLAs (e.g., LAPA), and primitive-guided VLAs (e.g., UniACT), LaDA consistently achieves strong results across task suites. Its performance on LIBERO-Long (86.4\%) further suggests that the proposed language-grounded, decoupled action representation helps capture shared motion semantics and supports compositional generalization in long-horizon control.

\begin{table}[t]
\centering
\caption{Comparison of success rates on LIBERO. LaDA achieves the highest overall performance and excels on long-horizon tasks, highlighting the benefit of language-grounded soft-label contrastive learning.}
\label{tab:libero_results}
\setlength{\tabcolsep}{3pt}
\renewcommand{\arraystretch}{1.1} 

\small
\begin{tabular}{lcccccc}

\toprule
\multirow{2}{*}{\textbf{Model}} & \multirow{2}{*}{\textbf{Params}} & \multicolumn{5}{c}{\textbf{LIBERO Success Rate (\%)}} \\
\cmidrule{3-7}
 & & \textbf{Spatial} & \textbf{Object} & \textbf{Goal} & \textbf{Long} & \textbf{Avg} \\
\midrule
UniACT \cite{zheng2025universal} & 0.5B & 65.0 & 78.0 & 68.0 & 47.0 & 64.5 \\
LAPA \cite{yelatent} & 7B & 73.8 & 74.6 & 58.8 & 55.4 & 65.7 \\
DP \cite{chi2023diffusion} & 147M & 78.3 & 92.5 & 68.3 & 50.5 & 72.4 \\
Octo \cite{ghosh2024octo} & 93M & 78.9 & 85.7 & 84.6 & 51.1 & 75.1 \\
OpenVLA \cite{kim2025openvla} & 7.5B & 84.7 & 88.4 & 79.2 & 53.7 & 76.5 \\
MDT \cite{reuss2024multimodal} & 380M & 78.5 & 87.5 & 73.5 & 64.8 & 76.1 \\
CLIP-RT* \cite{kang2024cliprt} & 0.6B & 86.8 & 77.2 & 86.8 & 59.4 & 77.5 \\
SpatialVLA \cite{qu2025spatialvla} & 4B & 88.2 & 89.9 & 78.6 & 55.5 & 78.1 \\
CoT-VLA \cite{zhao2025cot} & 7B & 87.5 & 91.6 & 87.6 & 69.0 & 81.1 \\
WorldVLA \cite{cen2025worldvla} & / & 87.6 & 96.2 & 83.4 & 60.0 & 81.8 \\
Dita \cite{hou2025dita} & 334M & 84.2 & 96.3 & 85.4 & 63.8 & 82.4 \\
ThinkAct \cite{huang2025thinkact} & 7B & 88.3 & 91.4 & 87.1 & 70.9 & 84.4 \\
$\pi$-FAST \cite{pertsch2025fast} & 2B & \textbf{96.4} & 96.8 & 88.6 & 60.2 & 85.5 \\
GR00T-N1.5 \cite{bjorck2025gr00t} & 3B & 92.0 & 92.0 & 86.0 & 76.0 & 86.5 \\
MolmoAct \cite{lee2025molmoact} & 7B & 87.0 & 95.4 & 87.6 & 77.2 & 86.6 \\
FlowVLA \cite{zhong2025flowvla} & 8.5B & 93.2 & 95.0 & 91.6 & 72.6 & 88.1 \\
CLIP-RT \cite{kang2024cliprt} & 1.3B & \underline{95.2} & \textbf{99.2} & \textbf{94.2} & \underline{83.8} & \underline{93.1} \\
\midrule
\rowcolor[HTML]{EFEFEF} 
\textbf{Ours (LaDA)} & \textbf{0.6B} & \underline{95.2} & \textbf{99.2} & \underline{93.6} & \textbf{86.4} & \textbf{93.6} \\
\bottomrule
\end{tabular}
\end{table}

\begin{table*}[t]
\centering
\caption{Simulation success rates of LaDA and baseline methods on the nine MimicGen tasks. 
We use the following abbreviations: C (Coffee), S (Stack), ST (StackThree), T (Threading), and TPA (ThreePieceAssembly), with D0 and D1 indicating different demonstration subsets. 
LaDA achieves the highest average performance, with notable gains on multi-step and long-horizon tasks.
}
\label{tab:mimicgen_results}
\small
\renewcommand{\arraystretch}{1.2} 
\begin{tabular}{lcccccccccc}
\hline
& \multicolumn{10}{c}{\textbf{MimicGen Simulation Success Rates}} \\ \cline{2-11} 
\multirow{-2}{*}{\textbf{Model}} & \textbf{C\_D0} & \textbf{C\_D1} & \textbf{S\_D0} & \textbf{S\_D1} & \textbf{ST\_D0} & \textbf{ST\_D1} & \textbf{T\_D0} & \textbf{TPA\_D0} & \textbf{TPA\_D1} & \textbf{Avg.} \\ \hline
OpenVLA \cite{kim2025openvla}                 & 42\% & 18\% & 84\% & 86\% & 36\% & 20\% & 20\% & 28\% & 8\%  & 38\% \\
Task-conditioned        & 66\% & 24\% & 88\% & 68\% & 30\% & 6\%  & 74\% & 20\% & 0\%  & 42\% \\
Subgoal-conditioned     & 76\% & 26\% & 88\% & 74\% & 24\% & 6\%  & 78\% & 20\% & 2\%  & 44\% \\
Motion-conditioned      & 68\% & 32\% & 92\% & 84\% & 38\% & 16\% & 58\% & 30\% & 4\%  & 47\% \\
Subgoal self-reflection & 80\% & 32\% & 88\% & 78\% & 32\% & 6\%  & 80\% & 34\% & 2\%  & 48\% \\
Phoenix \cite{xia2025phoenix}                 & 94\% & 48\% & 96\% & 86\% & 50\% & 20\% & 68\% & 52\% & 6\%  & 58\% \\
CLIP-RT* \cite{kang2024cliprt}       & 77\% & 34\% & 93\% & 87\% & 68\% & 52\% & 32\% & 11\% & 4\%  & 51\% \\ \hline
\rowcolor[HTML]{EFEFEF} 
\textbf{Ours}                    & 94\% & 46\% & 96\% & 95\% & 76\% & 71\% & 48\% & 50\% & 25\% & 67\% \\ \hline
\end{tabular}
\end{table*}

\subsubsection{Experiments on MimicGEN}
\noindent\textbf{Evaluation Protocol.}
To evaluate multi-task generalization in contact-rich manipulation settings, we further evaluate LaDA on MimicGen \cite{mandlekar2023mimicgen}, which generates large-scale demonstrations with diverse object interactions and initial state distributions. We evaluate nine manipulation tasks, each containing approximately 1,000 human demonstrations. For each task, 50 rollout trials are conducted, and the average success rate is reported to measure robustness across varying task conditions.

\noindent\textbf{Baselines.} We compare LaDA with representative prior methods: OpenVLA \cite{kim2025openvla} (end-to-end VLA), task-conditioned policies (task description conditioned, variant of RT-1 \cite{brohan2022rt} /Octo \cite{ghosh2024octo}), subgoal-conditioned policies (subgoals predicted by LLaVA-v1.5 \cite{liu2023visual}, similar to PaLM-E \cite{driess2023palm}), motion-conditioned policies (low-level motion instructions predicted by LLaVA-v1.5 \cite{oord2018representation}, variant of RT-H), subgoal self-reflection policies (subgoal-conditioned policy augmented with self-reflection), as well as Phoenix and a fine-tuned CLIP-RT*. 
We follow baseline configurations and training protocols from \cite{xia2025phoenix}, including task splits, task descriptions, preprocessing, and simulation environments, to ensure consistency and fair comparison. Detailed implementation and training settings for these baselines can be found in \cite{xia2025phoenix}.

\noindent\textbf{Results and Discussion.}
As summarized in \cref{tab:mimicgen_results}, LaDA achieves the highest average success rate across all nine MimicGen tasks, consistently outperforming all baselines. Notably, it performs particularly well on multi-step and long-horizon tasks such as StackThree\_D1, highlighting its robustness in handling temporally extended manipulation. These results suggest that LaDA’s language-guided soft-label contrastive learning effectively supports coherent control and facilitates generalization across diverse manipulation scenarios. Despite not incorporating any self-correction strategy, LaDA improves the average success rate by roughly 9\% over Phoenix and 16\% over CLIP-RT*, highlighting the effectiveness of its language-grounded semantic alignment.

\begin{figure}[t]
  \centering
  \includegraphics[width=\linewidth]{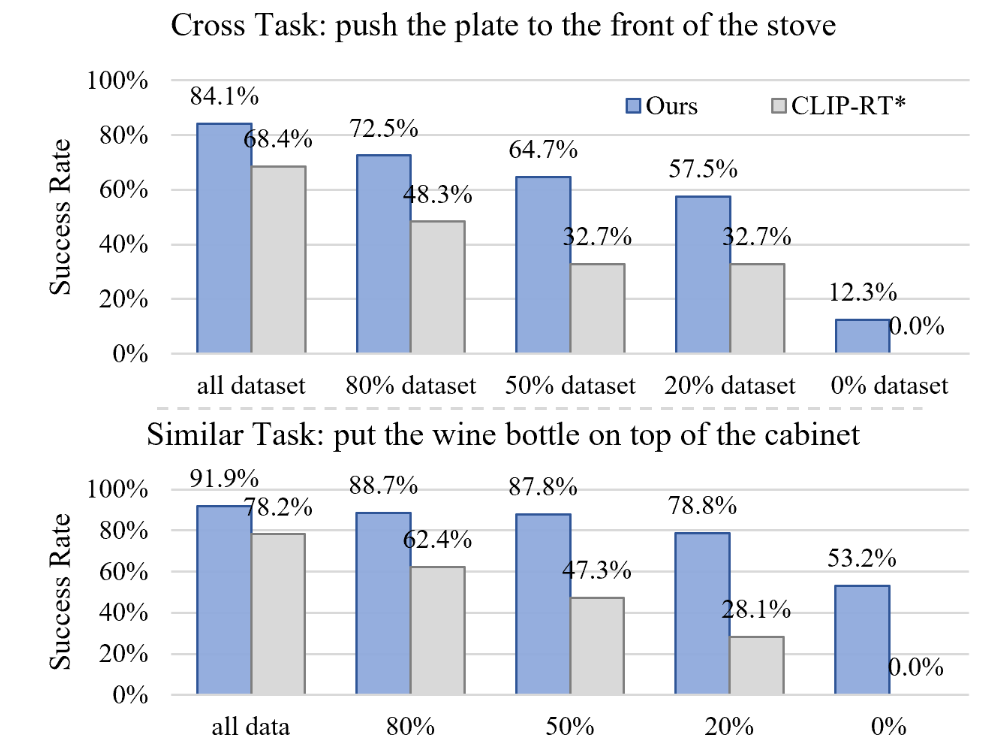}
   \caption{Generalization evaluation of LaDA on novel and semantically related tasks.
}
   \label{fig1.generalization-result}
\end{figure}

\subsubsection{Generalization Evaluation.}
We further evaluate LaDA’s generalization ability in the LIBERO-Goal benchmark under two challenging settings: (1) cross-task generalization, where instructions describe entirely new tasks not seen during training (e.g., “push the plate to the front of the stove”), and
(2) similar-task generalization, where instructions preserve the underlying semantics but shift the target goal (e.g., “put the wine bottle on top of the cabinet”). This evaluation tests whether LaDA can transfer shared motion semantics to tasks that differ either in intent (cross-task) or target configuration (similar-task). Success rates are averaged over four training data ratios (0\%, 20\%, 50\%, 100\%), each computed from 1,000 rollouts with 20 random seeds.

As shown in \cref{fig1.generalization-result}, LaDA consistently outperforms CLIP-RT* across both settings. For the cross-task “push” instruction, CLIP-RT* achieves 0\% success, while LaDA reaches 12.3\%, indicating that its language-grounded action representation enables the reuse of primitive-level semantics beyond the training distribution. Improvements on similar-task instructions further demonstrate that LaDA maintains coherent alignment across vision, language, and control when task goals shift but underlying motion structures remain related.

\begin{figure}[t]
  \centering
  \includegraphics[width=\linewidth]{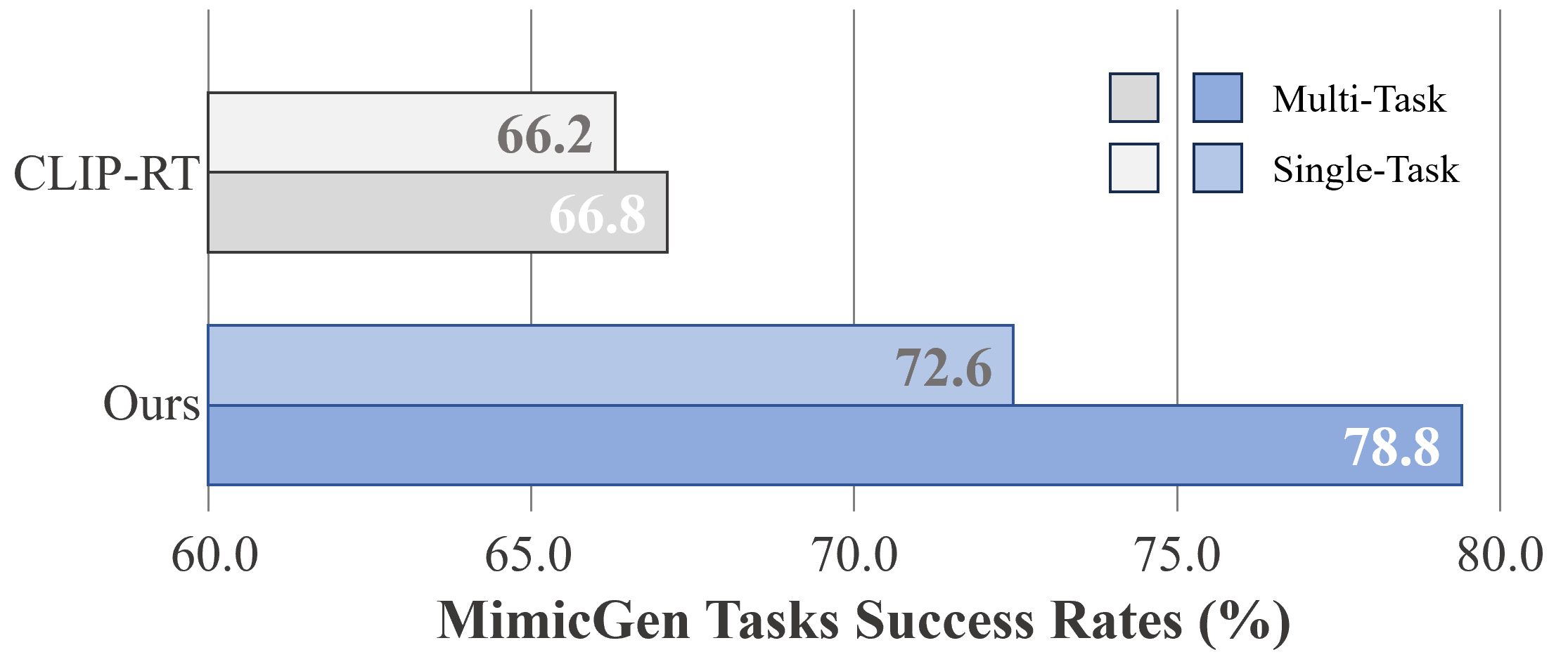}
   \caption{Comparison of average success rates on MimicGen tasks (Stack, StackThree, Threading) under single-task and multi-task training. LaDA consistently outperforms CLIP-RT, with larger gains in the multi-task setting.}
   \label{fig.mimicgen-generalization-result}
\end{figure}

We additionally investigate multi-task transfer in the MimicGen benchmark. As shown in \cref{fig.mimicgen-generalization-result}, LaDA benefits substantially from multi-task training, whereas CLIP-RT shows only marginal gains. This suggests that LaDA’s semantic structure facilitates more effective sharing of motion patterns across related manipulation skills.

\begin{table}[t]
\centering
\caption{Ablation study of LaDA components on the LIBERO benchmark.
Removing the proposed soft-label contrastive learning (SCL) or the adaptive weighting mechanism (AW) leads to clear performance degradation, verifying the importance of fine-grained semantic alignment and balanced optimization in LaDA. }
\label{tab:ablation_libero}
\small
\begin{tabular}{lccccc}
\toprule
\textbf{Methods} & \textbf{Spatial} & \textbf{Object} & \textbf{Goal} & \textbf{Long} & \textbf{Average} \\ 
\midrule
w/o SCL & 79.2 & 82.8 & 76.6 & 63.4 & 75.5 \\
w/o AW & 93.6 & 94.4 & 87.2 & 74.4 & 87.4 \\
\midrule
\rowcolor[HTML]{EFEFEF}
\textbf{LaDA} & \textbf{95.2} & \textbf{99.2} & \textbf{93.6} & \textbf{86.4} & \textbf{93.6} \\
\bottomrule
\end{tabular}
\end{table}

\begin{figure}[t]
  \centering
  \includegraphics[width=\linewidth]{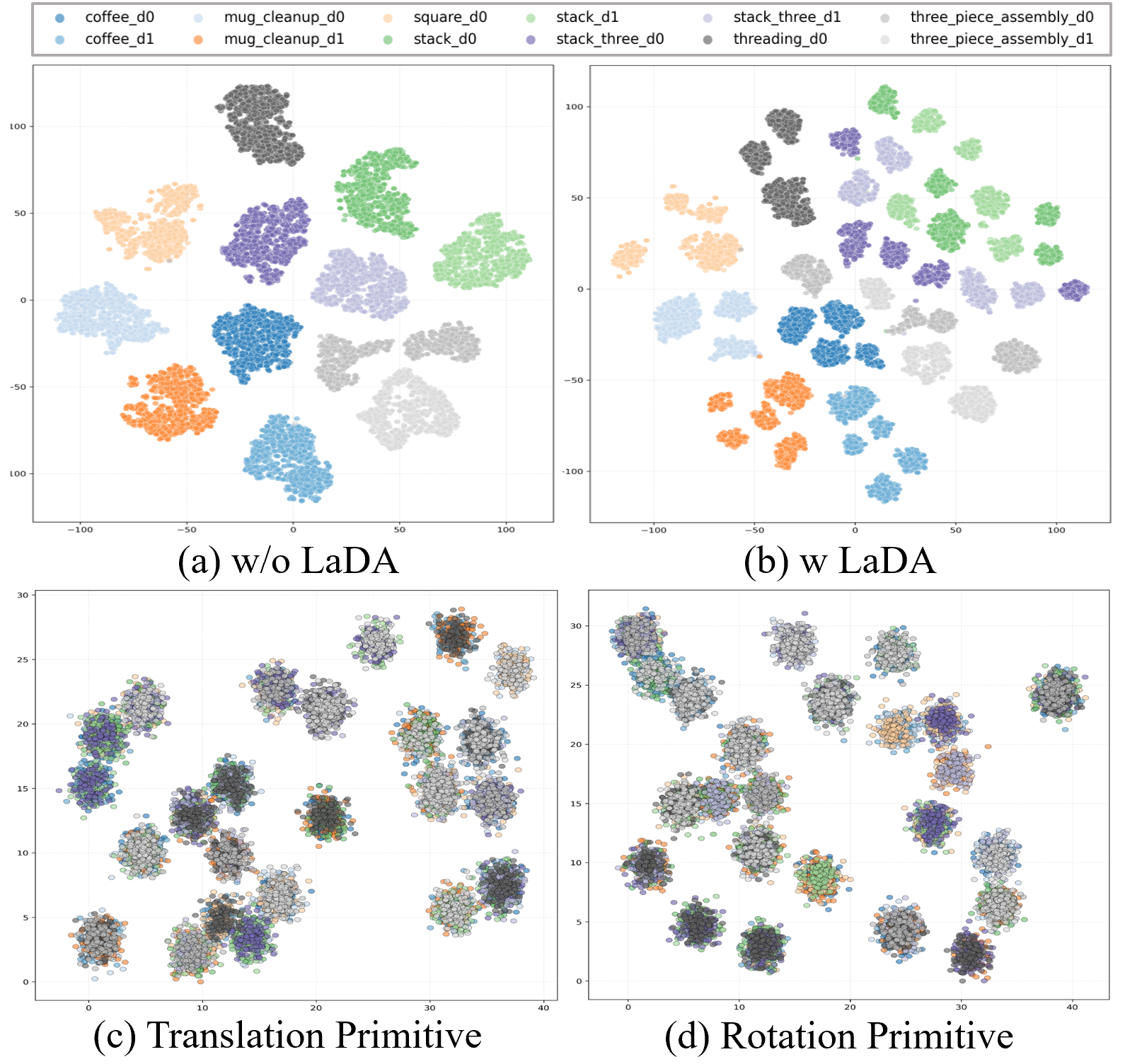}
   \caption{t-SNE visualization of learned action embeddings.
(a,b) Embedding distributions without and with LaDA, where LaDA yields more compact and semantically structured clusters.
(c,d) t-SNE projections of translation and rotation primitives, showing that actions from different tasks exhibit overlapping patterns, indicating consistent cross-task motion semantics.}
   \label{fig.tsne}
\end{figure}

Finally, we visualize learned action embeddings using t-SNE (\cref{fig.tsne}). Compared to the ablated variant without LaDA, the embeddings form clearer and more coherent clusters, reflecting improved organization of motion semantics. For both translation and rotation primitives, actions from different tasks display overlapping regions—evidence that LaDA captures reusable cross-task motion patterns and aligns them consistently across similar tasks.

\subsection{Real-World Experiments}
We deploy LaDA on a 7-DoF Franka Emika Panda using a static third-person RealSense D435i camera for a real-world pick-and-place task in which the robot grasps a cube and places it into a box. The model is pretrained as described in \cref{sec: pretraining} and fine-tuned with 100 human demonstrations.

\begin{figure}[t]
  \centering
  \includegraphics[width=\linewidth]{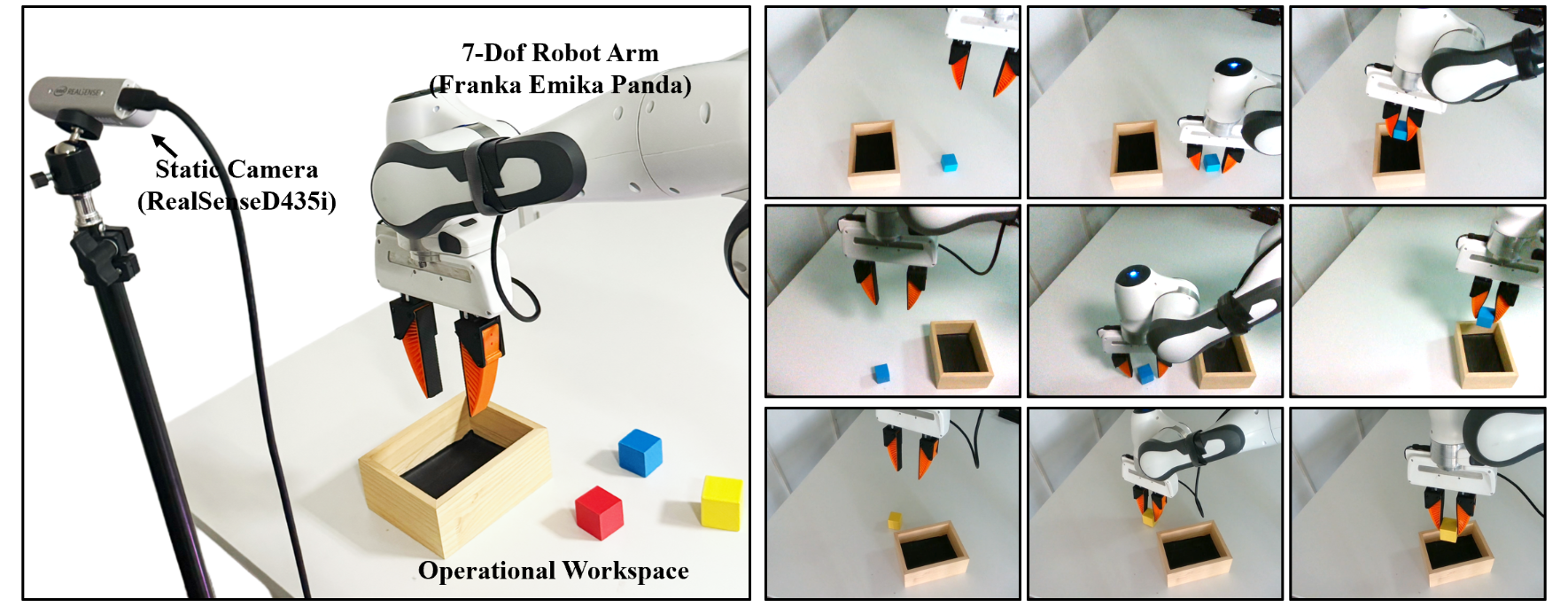}
   \caption{Real-world setup with a 7-DoF Franka Panda and a third-person static RGB camera (left). Representative rollout snapshots (right) show successful pick-and-place execution.}
   \label{fig.real-world}
\end{figure}

As shown in \cref{fig.real-world}, LaDA performs reliably on the physical robot and maintains robustness under variations in lighting, object pose, object color, and box placement. Across trials, the policy produces stable grasps and accurate placements, demonstrating that the learned representations transfer effectively to real-world execution conditions. 

\subsection{Ablation Studies}
We conduct ablation studies to evaluate the contribution of LaDA’s key components: the soft-label contrastive learning (SCL) strategy and the adaptive weighting mechanism (AW). Results on the LIBERO benchmark are summarized in \cref{tab:ablation_libero}. Removing the soft-label formulation and reverting to hard contrastive targets leads to a clear degradation across all task categories, indicating that fine-grained semantic alignment is crucial for capturing shared motion structure. Likewise, disabling the moving-average–based adaptive weighting results in lower and less consistent performance, reflecting the importance of maintaining balanced optimization between contrastive and imitation signals. Together, these ablations confirm that both components play complementary roles in stabilizing training and improving policy effectiveness.

\section{Conclusion}
We introduced LaDA, a language-guided framework that achieves fine-grained semantic alignment between visual observations, linguistic instructions, and robotic actions. By decomposing continuous 7-DoF control into interpretable, language-grounded primitives—translation, rotation, and gripper control—LaDA bridges the gap between high-level task semantics and low-level motor execution. A semantic-guided soft-label contrastive objective enables LaDA to align actions across tasks through graded linguistic affinities, facilitating efficient knowledge transfer and strong compositional generalization. An adaptive weighting mechanism further stabilizes joint optimization of imitation and contrastive objectives. Extensive evaluations on both simulated and real-world benchmarks, including LIBERO and MimicGEN, validate that LaDA delivers state-of-the-art performance and robust generalization to unseen instructions. In summary, LaDA demonstrates that language can serve as an effective semantic bridge for unifying perception and control, paving the way toward scalable, interpretable, and generalizable robotic learning systems.
\section{Acknowledgments}
This work was supported by the New Generation Artificial Intelligence-National Science and Technology Major Project (2025ZD0123003).

{
    \small
    \bibliographystyle{ieeenat_fullname}
    \bibliography{main}
}

\end{document}